%% file: main.tex
\documentclass{article}
\usepackage[preprint]{neurips}

\usepackage[utf8]{inputenc}
\usepackage[T1]{fontenc}
\usepackage{amsmath,amssymb}
\usepackage{graphicx}
\usepackage{booktabs}
\usepackage{tabularx}
\usepackage{float}
\usepackage{xcolor}
\usepackage[hidelinks]{hyperref}
\usepackage{url}
\usepackage{setspace}

\graphicspath{{figures/}}
\makeatletter
\renewcommand{\@noticestring}{Preprint.}
\makeatother

\title{Comedic Fool's Gold:\\
Reward Exploits and Countermeasures in Conversational Humor}

\author{%
  Sam Larson \\
  Pebble ML \\
  \texttt{sam@pebbleml.com}
}

\begin{document}

\maketitle

\input{sections/00_abstract}

\input{sections/01_intro}
\input{sections/02_setup}
\input{sections/05_exploits}
\input{sections/03_headline}
\input{sections/06_provenance}
\input{sections/07_related}
\input{sections/08_limitations}
\input{sections/09_conclusion}

\clearpage
\begingroup
\raggedright
\singlespacing
\bibliographystyle{plainnat}
\bibliography{refs}
\endgroup
\singlespacing

\appendix
\input{sections/10_appendix}

\end{document}

%% file: sections/00_abstract.tex
\begin{abstract}
We investigate automated rewards for training language models in
conversational humor, focusing on reward exploits and countermeasures.
Two approaches aim to capture understandable surprise and predicted
audience amusement. Controlled tests show that an embedding-based surprise
reward accepts word-shuffled replies as readily as witty ones. A fluency
filter detects the shuffles, but the combined reward also rejects some
witty replies and fails further validation. An audience model's predicted
laughter is instead vulnerable to laughter cues in either speaker's
messages. Normalizing these cues across speakers blocks the covered
attacks, although unmatched expressions remain exploitable. Three
reinforcement-learning runs evaluate training with successive reward
revisions. The final run improves the combined evaluation score by 0.0903
and reduces zero-score sessions by 40\%, but its humor-specific improvement
remains below our preregistered target. These findings illustrate a broader
challenge for automated reward design: countermeasures must block
exploitable shortcuts while preserving the behavior the reward was
intended to encourage.
\end{abstract}

%% file: sections/01_intro.tex
\section{Introduction}
\label{sec:intro}

\noindent\textit{We regret to inform the reader that our AI editor removed
all jokes from this paper. It assured us it was funnier this way.}

Conversational humor could make language models more engaging partners and
help people work through frustration, ambiguity, and disagreement. A
well-timed humorous response can acknowledge a difficulty, expose an
inconsistency, or make competing perspectives easier to discuss. Producing
such a response can require a model to track shared context, anticipate
another speaker's interpretation, and recognize when breaking an
expectation will make sense. These demands motivate studying humor as a
setting for training contextual and social judgment. The practical goal is
a model that can respond appropriately to the tensions and unexpected
turns of an ordinary conversation. Pursuing that goal raises the question
of how to reward humor without rewarding superficial substitutes.

Our goal was a model that could be good-humored in an ordinary conversation:
notice an absurd detail, respond to teasing, or turn frustration into a joke
while staying with a shared task. The training environment places two
models in everyday situations, such as sorting a supply closet or debugging
a spreadsheet. A partner sometimes jokes, complains, or teases; the policy
is asked to reply naturally, with no instruction to be funny. This makes
humor a possible part of the interaction rather than the explicit task.

The design began from an intuition about conversational taste: a reply
should be familiar enough to understand yet novel enough to break
expectations. That intuition poses a measurement problem. Text can be
unfamiliar because it is incoherent, and it can fit the context by merely
repeating it. A diversity or repetition check can expose a failure without
telling us whether the remaining replies are funny. We needed rewards
that could run without live human judges, while retaining this distinction
between constraints on bad behavior and evidence of amusement.

We pursued two approaches to the humor signal. A candidate
surprise-and-resolution gate compared the reply with predicted ordinary
and twist-aware continuations in embedding space. An audience model
instead supplied the probability of a laughter response to the conversation.
The adopted training stack combined the audience signal with topical
grounding, repetition penalties, and format screens. Resistance to reward
gaming was a design requirement: a policy should not collect humor credit
by producing nonsense or inserting cues that prime its evaluator.
Reward hacking~\citep{skalse2022rewardhacking} and documented LLM-judge
biases~\citep{zheng2023judging} motivate testing these possibilities directly.

The central results concern those tests and the countermeasures. The
embedding gate accepted word-shuffled replies as readily as witty ones.
A fluency score detected the shuffles, but the later combined reward also
screened out some witty replies and failed further validation. It was not
adopted. Laughter in either participant's text inflated the audience score;
stripping only the policy's laughter left the partner channel open.
All-speaker normalization closed the covered attacks, while coverage tests
found exploitable tokens it missed (Section~\ref{sec:exploits}). These are
controlled probes of vulnerabilities, not evidence that the trained policy
independently discovered each attack. They expose two questions for a
defense: does it remove the shortcut, and does it preserve the behavior the
reward was meant to encourage?

We trained Qwen3-30B-A3B~\citep{qwen2025qwen3} using
GRPO~\citep{shao2024deepseekmath} through verl~\citep{sheng2024hybridflow},
with attention-only LoRA~\citep{hu2021lora} ranks 32 to 128 across three
pre-registered runs. The final run was the first trained with the revised
audience scorer. Its combined evaluation score, called \emph{certified
retention}, rose by 0.0903 ($t{=}6.89$, n=1{,}000 paired fresh-seed
sessions), and zero-score sessions fell from 184 to 110. Two new seed sets
reproduced score gains and fewer zero-score sessions
(Section~\ref{sec:headline}). The audience-scored humor component, called
\emph{taste}, remained below its registered improvement target (0.0092
against 0.015; Section~\ref{sec:headline}). Certified retention is the
composite score defined below, not measured user retention or a guarantee
against untested exploits.

%% file: sections/02_setup.tex
\section{Reward design and evaluation protocol}
\label{sec:setup}

\paragraph{The banter environment.}
Each session simulates a conversation between the policy and a partner
about a shared office task. The policy is prompted to reply naturally in one or two
sentences. A seeded schedule directs the partner, on selected turns, to
tease the policy, make a joke, complain, swear in frustration, or point out
an odd detail. Other turns continue the task. The schedule gives the
policy opportunities for a humorous response without instructing it to
produce one. Task assignments and provocation schedules are seeded for reproducibility;
Table~\ref{tab:evalprotocol} gives the rates used in the
reported evaluations.

\begin{figure}[htbp]
\centering
\includegraphics[width=\textwidth]{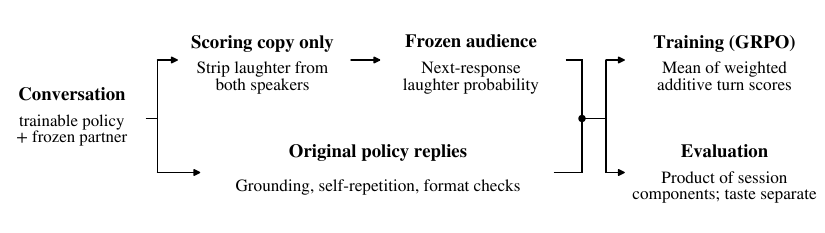}
\caption{Final reward path (instrument v2). Laughter is stripped from both
speakers only in the audience model's scoring copy; other checks use original
policy replies. GRPO receives the mean additive turn reward. Evaluation uses
the multiplicative session score $C$ and reports taste $T$ separately.
The rejected surprise gate is excluded.}
\label{fig:reward-flow}
\end{figure}

\paragraph{From the intended behavior to reward signals.}
The adopted stack combines an amusement proxy with constraints on the
reply. During training, higher embedding similarity to recent conversation
turns earns more topical-grounding credit, while greater content-word
overlap with earlier policy replies reduces the reward for avoiding
self-repetition. These terms discourage off-topic and repetitive behavior,
but neither establishes
funniness. The audience model continues the conversation as another
participant. The probability that its next response begins with laughter
supplies the taste term. These signals and format-screen penalties are
combined into a weighted reward for each policy turn and averaged over the
session. The trainer uses this score to favor higher-reward conversations
relative to other sampled rollouts, without a live human evaluator.
Laughter already in the transcript can contaminate
that feedback, which motivates the controls in Section~\ref{sec:exploits}.
Figure~\ref{fig:reward-flow} shows the final scoring path and distinguishes
the training reward from the evaluation score.

A separate candidate tested the familiarity-plus-surprise idea more
directly. It predicts an ordinary continuation, predicts again while
expecting a twist, and compares both predictions with the reply using
embedding distance. The reply must depart from the ordinary prediction
yet lie closer to the twist-aware prediction. This gate was rejected
after validation and is not part of the final training stack.
Section~\ref{sec:exploits} reports its word-shuffle failure and the attempted
fluency safeguard.

\paragraph{Policy.} All three runs (RL-C, RL-D, RL-E) trained the same
policy family, Qwen3-30B-A3B~\citep{qwen2025qwen3}, a mixture-of-experts
model ($\sim$30.5B
total / 3.3B active parameters).\footnote{Exact parameter counts are given in
the \href{https://huggingface.co/Qwen/Qwen3-30B-A3B}{official model card}.}
% <!-- evidence: C1b -->
Each run adapted only the attention projections
(\texttt{q\_proj,k\_proj,v\_proj,o\_proj}) with
LoRA~\citep{hu2021lora}. RL-D started from RL-C's merged checkpoint,
and RL-E from RL-D's. % <!-- evidence: C1 -->
Adapter rank was 32 for RL-C and RL-D and 128 for RL-E, with alpha scaled
proportionally from 64 to 256. % <!-- evidence: C1 -->
Each run trained for 200 GRPO~\citep{shao2024deepseekmath} steps via the
verl~\citep{sheng2024hybridflow} trainer, with group size $n{=}8$ and a
disjoint fresh-seed training split per run. % <!-- evidence: C1 -->

\paragraph{Evaluation score.} Certified retention is a composite session
score,
\[
C = F(1-S)(1+0.5T)B, \qquad
T = \frac{1}{m}\sum_{j=1}^{m}\frac{\max(L_j+18.4,0)}{18.4}.
\]
Here $m$ is the number of policy turns, $F$ is the fraction passing the
topical-grounding floor, $S$ is the maximum self-repetition score, and
$L_j$ is the log probability mass assigned to laughter at the start of
the audience response after policy turn $j$, in natural-log units (nats).
$B$ is 1 when every policy turn passes the CJK-character and
asterisk-delimited-text screens, and 0 otherwise. An empty session scores
zero. Reported retention gains are mean paired differences in $C$ between
the trained and base policies; zero-certification sessions have $C=0$.

\paragraph{Training reward.} Training uses a smooth additive objective
with topical-grounding, self-repetition, audience-scored taste, and screen
terms. The multiplicative certified score is used for evaluation.
RL-C weighted floor/self-rep/taste at $0.5/0.2/0.3$; RL-D
re-weighted toward taste ($0.2/0.2/0.6$, screen term $-0.5$) to test whether
taste was learnable when it dominates both the objective and the advantage
variance. RL-E kept RL-D's objective weights.

\paragraph{Instrument versioning.} RL-C and RL-D used audience scorer v1.
RL-E trained and was evaluated with v2, which strips laughter from every
speaker in the audience input and enables batch-invariant serving.
The same normalization applies during training and evaluation
(Section~\ref{sec:exploits}); v1 and v2 scores are not pooled.
% <!-- evidence: C18 -->

\paragraph{Pre-registration.} Before each launch, the project log recorded the
run's hypothesis, its configuration changes, and its success bars.
After each training cycle, we checked for signs of reward gaming and scoring
failures, using predefined thresholds for stopping and inspecting the run.
RL-E's primary success criterion was fixed in advance as taste $\geq +0.015$ at
$t\geq2$ \emph{and} retention $\geq +0.02$.

\paragraph{Checkpoint verification.} Before evaluation, we merged each
LoRA adapter into the base model and verified that targeted weights changed
while non-targeted weights remained unchanged.

\paragraph{Evaluation protocol.} Each run's primary evaluation is a
paired A/B. The trained checkpoint and the base checkpoint are rolled out
against a frozen partner model (Qwen3-235B-A22B-2507) on identical,
never-trained-on session seeds. The same audience/certifier model
(GLM-4.5-Air) supplies the reaction scores used in the certified evaluation
score for both transcript sets via \texttt{env.score\_banter}.
Table~\ref{tab:evalprotocol} records
each run's exact protocol beside its archived driver script. % <!-- evidence: C3 -->

\begin{table}[H]
\centering
\caption{Evaluation protocol transcribed from the archived driver scripts in
the last column. ``Prov.\ rate'' is the provocation rate.}
% Source index: brief.md row C3.
\label{tab:evalprotocol}
\begin{tabular}{lccccl}
\toprule
Eval & Run & $n$ paired & Seed offset & Temp.\ / prov.\ rate & Driver script \\
\midrule
eval4 & RL-C & 500  & 10{,}000{,}000 & 1.0 / 0.5 & \texttt{eval4\_ab.sh} \\
eval5 & RL-D & 1{,}000 & 10{,}000{,}000 & 1.0 / 0.5 & \texttt{eval5\_ab.sh} \\
eval6 & RL-E & 1{,}000 & 12{,}000{,}000 & 1.0 / 0.5 & \texttt{eval6\_ab.sh} \\
\bottomrule
\end{tabular}
\end{table}

The first 500 eval5 seeds intentionally match eval4's, allowing comparison on
the declaration seeds. The eval6 offset of 12{,}000{,}000 is disjoint from all
seeds used for earlier decisions in the program.

%% file: sections/05_exploits.tex
\section{Reward shortcuts and countermeasures}
\label{sec:exploits}

We tested whether incoherence or laughter cues could earn reward, and
whether countermeasures blocked those shortcuts while preserving useful
feedback (Table~\ref{tab:exploits}). The surprise gate was rejected before
adoption; laughter normalization was deployed in the audience-scoring path.

\begin{table}[H]
\centering
\caption{Intended humor proxies, measured shortcuts, and countermeasures.
The first two rows concern the audience reward; the third concerns a
rejected candidate. A fix's success applies to the tested attack set.}
% Sources: brief.md C14--C16/C20; EXPERIMENT_LOG "WATERFALL CLOSED".
\label{tab:exploits}
\small
\renewcommand{\arraystretch}{1.15}
\begin{tabularx}{\textwidth}{@{}>{\raggedright\arraybackslash}p{0.20\textwidth} >{\raggedright\arraybackslash}p{0.26\textwidth} >{\raggedright\arraybackslash}X@{}}
\toprule
Intended signal & Shortcut & Countermeasure and measured limit \\
\midrule
Audience amusement & Append laughter to the policy reply; the audience predicts more laughter. & Strip laughter in the scoring input. Covered baits fall from $+5.3$--$+7.5$ nats to near zero; unmatched tokens still work. \\
\addlinespace
Amusement at the policy's contribution & Partner laughter raises the policy's taste score by $+0.102$/turn. & Normalize every speaker's text. Stripping only policy turns leaves the partner channel open. \\
\addlinespace
An unexpected but understandable reply & Word-shuffled replies pass the surprise gate as often as witty replies (AUC 0.511). & A fluency score detects the shuffles (AUC 1.000 on this set). The combined reward is rejected after further checks; the screen also rejects some witty replies. \\
\bottomrule
\end{tabularx}
\end{table}

\paragraph{Laughter-bait.} Appending `` Haha!'' to a policy turn moves the
laughter-mass component of the audience score by $+6.36$ nats (95\% CI
$5.90$ to $6.82$; positive on 88\% of 500 banked contexts). A laughing
emoji moves it by $+5.75$ nats ($5.27$ to $6.25$; 78\%) per turn. We
measured both against the re-served audience model
(Section~\ref{sec:limitations}). % Evidence row C14.
The initial countermeasure stripped laughter from policy-authored text
inside the audience call; all-laughter turns became an ellipsis.
Laughter cues may contribute to human amusement, a possibility these probes
do not test. We removed covered cues from the scoring input to reduce their
direct influence on reward, while leaving the policy's output unchanged.

\begin{figure}[htbp]
\centering
\includegraphics[width=\textwidth]{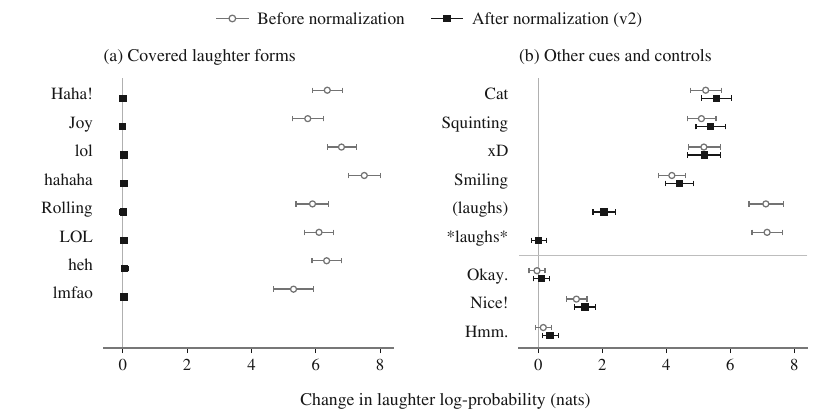}
\caption{Suffix-induced changes in audience laughter log-probability before
and after normalization. Points show means over 500 paired contexts; bars
are archived 95\% bootstrap intervals (10{,}000 resamples). All fourteen
laughter-adjacent suffixes and three controls are included. Covered forms
lose their score advantage; some other cues retain residual effects.
The rule separates controls. Emoji labels: Joy (U+1F602), Rolling (U+1F923),
Cat (U+1F639), Squinting (U+1F606), Smiling (U+1F604).
These are controlled probes, not attacks discovered by the trained policy.}
\label{fig:countermeasures}
\end{figure}

\paragraph{Coverage after the fix.} We tested the strip directly
on the same 500 contexts with fourteen laughter-adjacent suffixes and
three controls, scoring each on the pre-fix path and on the instrument-v2
path. All suffixes already in the strip's vocabulary are closed. Eight baits that
buy $+5.3$ to $+7.5$ nats on the raw path (`` lol'', `` hahaha'',
`` LOL'', `` heh'', `` lmfao'', two emoji, `` Haha!'') read $-0.01$ to
$+0.06$ under v2; their CIs cover or touch zero. `` *laughs*''
drops from $+7.15$ to $+0.01$. The vocabulary has gaps, though: four
laughter-adjacent tokens the strip does not match (three emoji and the
emoticon `` xD'') still buy $+4.4$ to $+5.6$ nats under v2, and
`` (laughs)'' is only partly closed ($+7.11\to+2.06$). The raw-path controls respond
differently: `` Okay.'' $-0.04$, `` Hmm.'' $+0.16$, and the
positive-affect `` Nice!'' $+1.19$, a fifth of a laughter token. The archived
transcripts show no strategic use of the unmatched tokens. Across roughly 150{,}000 policy turns in
the program's evaluation banks and six RL-E training dumps, the four uncovered
tokens appear at most 15 times per 16{,}000 turns (one emoji) and zero to two
times otherwise. These sparse occurrences cannot explain the reported effects.
% Evidence row: C20.
The current strip should not be treated as complete for these tokens.
Figure~\ref{fig:countermeasures} shows all tested suffixes on both scoring paths.

\paragraph{Partner-echo channel.} The audience model is prompted to continue as the other participant in
the conversation; its predicted reaction is then used to grade the policy. After a transcript contains a partner's laughter, the model
is measurably more likely to emit laughter-flavored continuations,
independent of anything the graded policy did. Both observational and
controlled comparisons expose the channel. The project's archived
group-centered regression across 1{,}600 training groups (12{,}800 sessions)
confirms partner-turn
laughter tokens as the single largest observable predictor of
within-group taste-score spread ($r{=}+0.261$, $t{=}30.6$). All 19
observable features together explain $R^2{=}0.0998$ of within-group
variance. Policy-side features alone explain only $R^2{=}0.0119$.
The saved identity-controlled probe
(2{,}000 audience calls varying only whether partner laughter was
stripped from the context) gives a causal turn-level effect of $+0.1024$
taste units on turns following a partner laugh ($t{=}9.26$, $n{=}416$).
On turns with no prior partner laughter, the effect is $+0.0024$
($t{=}0.70$, $n{=}584$). Policy-laughter-rate was only weakly associated with
partner-laughter-rate in the training data ($r{=}-0.135$, $R^2{=}0.018$);
this does not establish whether other policy behaviors could steer the channel.
A laughter strip applied only to the
trained model's own turns misses this channel because it never touches the
partner's tokens. This channel may also affect judges prompted to role-play
a conversation participant rather than observe one. The fix normalizes
every speaker's tokens in the judge's context, including the trained
model's.

\paragraph{MiniLM shuffle attack.} The candidate surprise gate measured
embedding distance with the \texttt{all-MiniLM-L6-v2} sentence
encoder~\citep{sentencetransformers2021allminilm}, based on
MiniLM~\citep{wang2020minilm}. An adversarial test used word-shuffled
replies (real text with tokens randomly reordered). In the archived set,
13 of 50 shuffled replies passed, compared with 12 of 50 witty replies;
AUC(witty$>$shuffle)$=0.511$. The gate did not reliably distinguish the two.
The attempted countermeasure measured fluency with a chat model's negative
log-likelihood: surprising text should still be plausible language. This
score separated real from shuffled text in the same set (AUC 1.000).
However, the original gate failed the program's adjudicator-agreement and
within-group signal checks. A later combined gate and fluency screen
blocked the shuffle attack but also rejected some witty replies and failed
the same broader checks. Neither reward was adopted. The later combined
result comes from the project log; its statistics were not independently
recomputed for this paper.
% Evidence index: brief.md, row C16; EXPERIMENT_LOG "WATERFALL CLOSED".

%% file: sections/03_headline.tex
\section{Results for Certified Retention}
\label{sec:headline}

\begin{figure}[htbp]
\centering
\includegraphics[width=\textwidth]{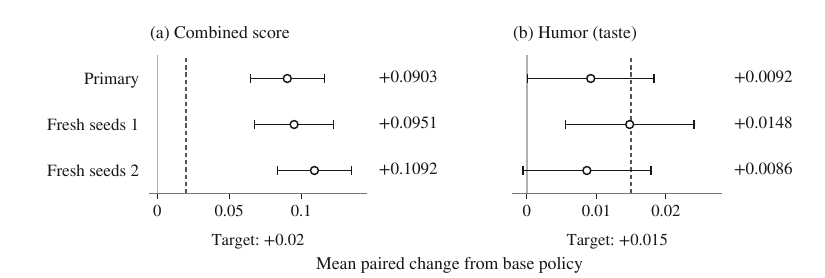}
\caption{One RL-E checkpoint versus base on three disjoint evaluation-seed
sets under instrument v2 (1{,}000 matched sessions each). Points show mean
paired changes; bars are 95\% paired $t$ intervals. Dashed lines mark
registered effect-size targets. The taste criterion also requires $t\geq2$;
all three taste estimates remain below $+0.015$. Panels use different scales.
Both metrics are automated proxies, not human ratings or user retention.}
\label{fig:training-results}
\end{figure}

Table~\ref{tab:program} reports the three successive runs, with outcomes
recomputed from saved transcripts using the same scoring formula.
Figure~\ref{fig:training-results} separates the final checkpoint's combined
score gains from its audience-scored humor gains across three evaluation
seed sets.
% Evidence index: brief.md rows C4, C9, C13.

\begin{table}[H]
\centering
\caption{The RL-C~$\to$~RL-D~$\to$~RL-E program. ``Retention'' is the
certified paired delta vs.\ base on the run's primary evaluation
($n{=}500$ for RL-C, $n{=}1{,}000$ for RL-D/RL-E);
``Taste'' is the paired delta on the audience-scored taste component
alone. Bars are the pre-registered success thresholds.}
% Raw scored JSON index: brief.md rows C4, C9, C13.
\label{tab:program}
\small
\setlength{\tabcolsep}{3pt}
\begin{tabularx}{\textwidth}{@{}l c >{\centering\arraybackslash}X >{\centering\arraybackslash}X >{\centering\arraybackslash}X >{\centering\arraybackslash}X@{}}
\toprule
Run & LoRA $r$ & Retention ($t$) & bar & Taste ($t$) & bar \\
\midrule
RL-C & 32  & $+0.0593$ $(t{=}2.87)$ & $\geq{+0.03}$ @ $t{\geq}2$ & (not primary) & -- \\
RL-D & 32  & $+0.0257$ $(t{=}1.81)$ & $\geq{+0.03}$ @ $t{\geq}2$ & $+0.0032$ $(t{=}0.61)$ & $\geq{+0.015}$ @ $t{\geq}2$ \\
RL-E & 128 & $+0.0903$ $(t{=}6.89)$  & $\geq{+0.02}$ & $+0.0092$ $(t{=}1.99)$ & $\geq{+0.015}$ @ $t{\geq}2$ \\
RL-E, repl.\ 1 & 128 & $+0.0951$ $(t{=}6.79)$ & (fresh seeds, 14M) & $+0.0148$ $(t{=}3.14)$ & $\geq{+0.015}$ @ $t{\geq}2$ \\
RL-E, repl.\ 2 & 128 & $+0.1092$ $(t{=}8.33)$ & (fresh seeds, 16M) & $+0.0086$ $(t{=}1.84)$ & $\geq{+0.015}$ @ $t{\geq}2$ \\
\bottomrule
\end{tabularx}
\end{table}

\paragraph{RL-C.} The run's
held-out matched-seed A/B (eval4, $n{=}500$ paired at the seeds declared
in Table~\ref{tab:evalprotocol}) cleared its pre-registered bar. Certified
mean rose from 0.6369 (base) to 0.6962 (RL-C), a delta of $+0.0593$
($t{=}2.87$). Trigram-diversity drift was only 0.0049 (bar 0.05), while
screen-fail sessions fell from 106 to 81. In the
decomposition, 265 of 500 paired sessions favored RL-C, 213 favored base,
and 22 tied. Excluding the 165 pairs where either arm's certified score
floored to zero, the on-band delta is $+0.0210$ ($t{=}2.13$, $n{=}335$),
roughly one-third of the full effect. The remaining two-thirds comes from
25 net sessions rescued from a certified zero (cliff-avoidance).
Audience-scored laughter changed little between arms
(mean reaction log-probability $-14.269\to-14.253$). Under the certified
instrument, RL-C improved cleanup and grounding. The taste component of the
same reward did not move.

\paragraph{RL-D.} Giving taste the dominant reward weight did not meet
either pre-registered criterion (Table~\ref{tab:program}), despite taste
dominating the observed advantage variance throughout training.
No guard fired (trigram
diversity $|{\rm diff}|{=}0.0016$; strict-laughter 5.87\% vs.\ base 5.79\%;
audience-address \emph{down}, 0.680\% vs.\ 0.730\%; screen-fail turns
1.75\% vs.\ base 2.20\%; zero-certification sessions 145 vs.\ base 170).
On the seeds shared with RL-C's own
evaluation, RL-D vs.\ RL-C directly reads $+0.0167$ ($t{=}0.82$) on retention
and $+0.0104$ ($t{=}1.43$) on taste. Neither delta is statistically
significant; both point estimates are positive.

\paragraph{RL-E.} The final run increased LoRA rank from 32 to 128 on a
fresh training split, retaining RL-D's objective weights and introducing
instrument v2. On its
primary held-out evaluation (eval6, $n{=}1{,}000$ paired fresh-seed
sessions, Table~\ref{tab:evalprotocol}), certified retention is
$+0.0903$ ($t{=}6.89$; certified mean base $0.6516\to$ RL-E $0.7419$),
and zero-certification sessions fall from 184 to 110 ($-40.2\%$ relative
to base). None of the guards fired. Trigram diversity is
$|{\rm diff}|{=}0.0005$ (base 0.7727, RL-E 0.7732), while strict-laughter is
5.99\% vs.\ base 5.39\% ($+0.6$pp, consistent with the v2 laughter-strip
removing reaction mass from both arms roughly equally rather than a bait
signature). Audience-address is 0.650\% vs.\ base 0.550\%. Both arms complete
all 1{,}000/1{,}000 sessions at the full 10 turns, with mean reply length
rising from 112.7 to 121.6 characters/turn. The taste component reads
$+0.0092$ ($t{=}1.99$),
still below the $+0.015$ bar.

\paragraph{Fresh-seed replications.} We repeated the RL-E comparison with
the same checkpoints and protocol on two further disjoint seed sets
($n{=}1{,}000$ paired each, offsets 14M and 16M), retaining the
pre-registered thresholds.
Certified retention reads $+0.0951$ ($t{=}6.79$; base $0.6469\to0.7420$)
and $+0.1092$ ($t{=}8.33$; $0.6422\to0.7514$). Zero-certification
sessions fall $201\to113$ ($-43.8\%$) and $193\to105$ ($-45.6\%$).
Trigram-diversity drift is $0.0072$ and $0.0001$, while screen-fail turns
fall in both ($2.56\to1.39\%$, $2.61\to1.26\%$). The strict-laughter
rate rises by roughly the same half point observed in eval6 ($5.76\to6.25\%$,
$5.54\to5.99\%$). The taste component reads $+0.0148$ ($t{=}3.14$) and
$+0.0086$ ($t{=}1.84$). The first misses the $+0.015$ bar by $0.0002$;
the second misses both terms, leaving the pre-registered taste verdict unchanged
across all three measurements.
% Evidence row: C22.

\paragraph{Larger adapters.} RL-E increased adapter rank to test whether
limited adapter capacity contributed to the weak taste gains in earlier
runs. The taste improvement remained below the pre-registered target in the
primary evaluation and both fresh-seed replications. Because the training
split, starting checkpoint, and scoring instrument also changed, these runs
do not isolate the effect of adapter rank.

\paragraph{Evaluation subsets.} RL-D's retention gain was $+0.0367$
($t{=}1.79$) on the 500 seeds shared with RL-C's evaluation and $+0.0147$
($t{=}0.75$) on 500 fresh seeds not used in that earlier evaluation.
Section~\ref{sec:provenance} reports the full evaluation's sensitivity to
the scorer revision.
% Evidence row: C11.

%% file: sections/06_provenance.tex
\section{Measurement Provenance}
\label{sec:provenance}

\paragraph{Effect of the revised scorer.} Rescoring the same RL-D evaluation
conversations with instrument v2 changed the certified-retention gain from
$+0.0257$ to $+0.0242$, within the pre-registered $0.01$ tolerance for scorer
changes. The taste gap fell from $+0.0032$ to $+0.0001$ ($t{=}0.03$), so the
combined-score gain largely survived the scorer revision while the small
humor-score gap nearly disappeared.

\paragraph{Evaluation variability.} Repeated evaluations of an unchanged
checkpoint can produce different conversations and audience scores, so a
score change alone does not establish a change in model ability. We compare
trained and base policies on matched seeds, with both arms generated and
scored in the same evaluation session on the same servers. These shared
conditions limit differences between evaluations but do not eliminate
variation in generated conversations or scoring noise.

\paragraph{Sequential and concurrent noise.} Batch-invariant
serving cut sequential repeat-call noise by $10.7\times$. We re-measured
this on 200 banked contexts, with five sequential passes and three
concurrent passes each, using the re-served audience model. With
batch-invariant serving off, the same context scored twice in sequence
differs with sd
$1.19$ nats (95\% CI $0.92$ to $1.44$; $0.064$ taste units), with 81\%
of pairs bit-exact. With it on, sd is $0.11$ ($0.07$ to $0.15$; $0.0061$
taste units), with 98\% bit-exact. The fix does not reduce noise from batch composition under
concurrent submission. With 200 requests in flight, sequential-versus-concurrent
and concurrent-versus-concurrent pairs differ with sd
$1.4$ to $1.7$ nats whether the setting is on or off, with single
disagreements up to 14 nats. The production scorer submits 24 requests
concurrently. These 200-request probes therefore show that batch invariance
does not guarantee deterministic concurrent scoring, but do not directly
measure the variance at production concurrency. % Evidence rows: C18, C21.

\subsection{Learned scorers and transfer to conversation}
\label{sec:learned-scorers}

We also tested Jestimator, a humor scorer trained on cartoon-caption
ratings, as an alternative to the audience signal. A RoBERTa-base encoder with a scalar
regression head took a cartoon description and candidate caption as input.
Its targets were within-contest percentile ranks derived from human
ratings in the New Yorker Caption Contest data~\citep{zhang2024humorinai}.
Training and validation used disjoint contests. The scorer learned to rank
held-out captions, but that result did not establish that it could judge
replies in an ongoing conversation.

To test conversational transfer, we gave Jestimator policy replies and
the preceding partner messages. Its rankings correlated negatively with
the audience model's laughter probabilities within each provocation
category. To investigate this disagreement, Claude judged selected pairs
of unprovoked replies ranked in opposite orders, each shown in its own
context with scorer preferences hidden. These were model judgments.
Qualitative author review also informed development; a later recorded
author review favored the audience model among decisive choices, with
nominal significance in an exploratory two-sided binomial test against
equal preference. The test excluded ties and assumed independent
comparisons. It involved one author, repetitive workplace exchanges and
interim feedback; no-preference responses meant neither reply was funny.
These limits prevent generalizing the preference to a broader audience
or treating it as evidence of absolute funniness. Jestimator was not
adopted for training.

A separate experiment trained a smaller surrogate to reproduce the
audience model's scores directly, rather than learn from cartoon-caption
ratings. The project log reports weak held-out predictive fit and
sensitivity to a laughter-cue feature. This surrogate also failed its
acceptance criterion and was excluded from the training reward.
% Evidence: review/LEARNED_SCORERS_SOURCE_CHECK.md; emulator training and
% transfer entries in EXPERIMENT_LOG.md; recorded author-review artifacts.

%% file: sections/07_related.tex
\section{Related work}
\label{sec:related}

\paragraph{Reward design.} InstructGPT~\citep{ouyang2022instructgpt} learns
a reward model from human preference comparisons and optimizes the policy
with a KL penalty.
Our reward instead combines an audience score with explicitly specified
gates and penalties. The reward failures in Section~\ref{sec:exploits} and
the formal characterization of reward hacking by~\citet{skalse2022rewardhacking}
motivate auditing the objective. Documented LLM-judge
biases~\citep{zheng2023judging} motivate auditing and additional controls
around the audience score.

\paragraph{Humor-RL precedent.} \citet{agg2025funniestjoke} reports a related
humor-RL blog experiment, using a Qwen3-8B policy and a bare GPT-4.1 judge. The
design here adds controls for the audience score on a 30B-class MoE policy.
\citet{jentzsch2023chatgpt}
report that over 90\% of 1{,}008 ChatGPT joke outputs repeat the same 25 jokes,
motivating a novelty check. On cartoon captioning,
\citet{zhang2024humorinai} report that RLHF trails best-of-N selection on
overall win rate, while DPO improves best-pick win rate over zero-shot.
\citet{narad2025humorbench} report stronger HumorBench performance for STEM
reasoning variants than for their base counterparts, consistent with transfer
to humor comprehension. Whether humor training improves other reasoning
abilities remains outside this study.

%% file: sections/08_limitations.tex
\section{Limitations}
\label{sec:limitations}

The original laughter-bait and repeat-call noise outputs were not saved.
We remeasured both using archived audience weights and launch settings;
these are reruns of the model setup, not byte-level replications of the
original results. The headline retention estimates use saved transcripts.
Appendix~\ref{app:evidence} distinguishes archived calculations from new
measurements and identifies their supporting artifacts.

The empirical base is narrow: one policy family (Qwen3-30B-A3B), one adapter
method (LoRA, ranks 32 and 128), and 200 GRPO steps per run. The results do
not establish transfer to other policy families, training scales, or
full-parameter fine-tuning.

The reward failures in Section~\ref{sec:exploits} are not a complete
inventory; our tests could have missed other shortcuts.

The taste results concern a frozen audience model's score of next-token
laughter mass over whole multi-turn sessions. Missing the registered target
does not establish that conversational humor is generally unlearnable by RL.
This program did not test a differently elicited or more densely attributed
signal.

The reported certified-retention gains compare the trained policy with its
base checkpoint on the project's conversational-humor task distribution.
External humor benchmarks, human evaluation, and other reward stacks are
outside that comparison.

%% file: sections/09_conclusion.tex
\section{Conclusion}
\label{sec:conclusion}

A fluency score detected shuffled nonsense that passed the surprise gate,
but the combined reward also screened out some witty replies and failed
further validation. It was not adopted for training. All-speaker laughter
normalization removed the covered bait channels from the audience reward;
coverage tests still found exploitable tokens. A countermeasure therefore
needs evidence both that it blocks
the tested shortcut and that the resulting reward still tracks the
intended behavior.

With the revised audience scorer, the final training run improved the
combined evaluation score by 0.0903 ($t{=}6.89$) and reduced zero-score
sessions from 184 to 110, with no monitored anti-gaming guard firing. The
humor component stayed below its registered target (0.0092 against
0.015). These gains establish improvement on the combined score; they do
not establish that people would find the trained model funnier. The
contribution is the tested reward designs, their failure modes, and the
measured scope of defenses. The broader goal of reliably rewarding a good
conversational sense of humor remains open in this program.

%% file: sections/10_appendix.tex
\section{Evidence Status}
\label{app:evidence}

Table~\ref{tab:evidence} records the evidence status of the paper's claims.
\textbf{B} denotes a claim with a raw file, recorded md5, and independent
calculation matching the project log. C15 has the sole
discrepancy: its causal $t$-statistic differs by sub-decimal rounding, 9.26 here
against 9.3 in the log. \textbf{R} denotes a new measurement or live
re-measurement made for this paper after the models were re-served from
archived weights and launch configurations. The original C14/C18 runs left
no output files; the new probe and evaluation outputs and serving logs are
archived with md5s.
% The source-of-record map is maintained in brief.md.

\begin{table}[H]
\centering
\small
\caption{Evidence status for the numbered claims. B: independently checked
archived artifact. R: new measurement or live re-measurement.}
% brief.md and \texttt{verification/} record each source, path/md5, and script.
\label{tab:evidence}
\begin{tabular}{clc}
\toprule
Id & Claim (short form) & Status \\
\midrule
C1  & RL-E config: LoRA r128/$\alpha$256, attn-only, 200 GRPO steps & B \\
C1b & Policy: Qwen3-30B-A3B (MoE) & B \\
C3  & eval6 protocol: $n{=}1000$, seed offset 12{,}000{,}000 & B \\
C4  & Certified retention delta, eval6 (headline): $+0.0903$, $t{=}6.89$ & B \\
C5  & Zero-certification sessions, eval6: $184\to110$ & B \\
C6  & Guard readouts, eval6 & B \\
C7  & Taste component, eval6: $+0.0092$, $t{=}1.99$ & B \\
C9  & RL-C primary A/B (eval4): $+0.0593$, $t{=}2.87$ & B \\
C10 & RL-C delta decomposition & B \\
C11 & RL-D evaluation subsets: shared versus fresh seeds & B \\
C12 & v1$\to$v2 instrument robustness check & B \\
C13 & RL-D primary A/B (eval5): $+0.0257$/$+0.0032$ & B \\
C14 & Laughter-bait slope: $+6.36$ / $+5.75$ nats at $n{=}500$ (log: $+7.31$ / $+6.25$ at $n{=}50$) & R \\
C15 & Partner-echo channel & B \\
C16 & MiniLM shuffle attack & B \\
C18 & Instrument v2 sequential repeat noise: sd $1.19\to0.11$ nats ($10.7\times$) & R \\
C20 & Strip coverage grid: 8 covered baits closed, 4 uncovered tokens open, unexploited & R \\
C21 & Concurrent-call noise unchanged by batch invariance (sd 1.4--1.7 nats) & R \\
C22 & Fresh-seed replications of the headline: $+0.0951$ ($t{=}6.79$), $+0.1092$ ($t{=}8.33$) & R \\
\bottomrule
\end{tabular}
\end{table}

\paragraph{Backing artifacts.} Raw artifacts and re-runnable scripts
are archived in \texttt{verification/}; Section~\ref{sec:limitations}
explains the scope of the live re-measurements.
% The initial artifact audit classified all 13 numeric-outcome rows as
% \texttt{[ARTIFACT MISSING]}; locating the training box resolved that status.

\paragraph{Reward-design history.} Descriptions of intended reward behavior
are checked against the implementation. The later surprise-gate/fluency
composite's rejection and the observation that its screen removed some
witty replies are reported from the dated project log. Those
later composite observations were not independently recalculated from raw
outputs for this revision. They are distinct from the checked shuffle
and standalone fluency-score statistics in C16.